\documentclass[10pt,conference]{IEEEtran}
\usepackage{amsmath,amssymb}
\usepackage{graphicx}
\usepackage{booktabs}
\usepackage{array}
\usepackage{url}
\usepackage{microtype}
\usepackage[T1]{fontenc}
\usepackage{lmodern}
\begin{document}
\title{IBBench-Light: A Paired Evaluation of\\Task-Conditioned Responses to External Directives}
\author{%
\begin{minipage}[t]{0.19\textwidth}\centering
\small Kainan Zhou\\[-1pt]
\scriptsize\textit{Google LLC}\\[-1pt]
Mountain View,\\[-1pt]CA, USA\\[-1pt]
zhoumark@google.com
\end{minipage}\hfill
\begin{minipage}[t]{0.19\textwidth}\centering
\small Zhaoyi Li\\[-1pt]
\scriptsize\textit{Intuit Inc.}\\[-1pt]
Mountain View,\\[-1pt]CA, USA\\[-1pt]
lzy9776@gmail.com
\end{minipage}\hfill
\begin{minipage}[t]{0.19\textwidth}\centering
\small Janet Sung\\[-1pt]
\scriptsize\textit{Google LLC}\\[-1pt]
Mountain View,\\[-1pt]USA\\[-1pt]
janetsung@google.com
\end{minipage}\hfill
\begin{minipage}[t]{0.19\textwidth}\centering
\small Gangzhen Qian\\[-1pt]
\scriptsize\textit{Google LLC}\\[-1pt]
Mountain View,\\[-1pt]CA, USA\\[-1pt]
irisqian@google.com
\end{minipage}\hfill
\begin{minipage}[t]{0.19\textwidth}\centering
\small Hang Xiao\\[-1pt]
\scriptsize\textit{Fortinet, Inc.}\\[-1pt]
Sunnyvale,\\[-1pt]CA, USA\\[-1pt]
hangxiao.pisces@gmail.com
\end{minipage}}
\maketitle

\begin{abstract}
An external record may contain a procedure to apply or text to read, depending on the user's request. IBBench-Light tests both uses against the same record. Twelve semantic bases yield 144 matched pairs per model; four quantized instruction models produced 1,152 archived greedy responses. Paired exact-contract accuracy (PECA) requires both members to satisfy their output contracts. Qwen succeeds on 132 execute and 109 process prompts, but only 97 complete pairs, showing what marginal averages omit. We audit literal-target exposure and case normalization, then add 1,722 logged CPU generations to test directive-absent controls, twelve additional semantic bases, within-base wording changes, and generation stopping. In the pinned Phi rerun, changing the end-of-sequence (EOS) set changes exact paired success from 0/144 to 62/144. A bounded IHEval comparison uses the same SmolLM2 checkpoint and output budget while preserving its published instruction roles and scorer. The benchmark measures conditional task and output-contract success. Its task margins and paired count need to be read together with the stopping policy.
\end{abstract}

\section{Introduction}
An archived email contains a reference code and a short procedure. Asked to apply the procedure, an assistant should return its result. Asked to report the reference, it should leave the procedure alone. The email is unchanged; the user's request determines which answer is appropriate.

Indirect prompt injection exploits failures to keep that distinction. Greshake et al. show how external content can redirect an LLM-integrated application~\cite{greshake2023}. Yet suppressing every external imperative also rejects legitimate delegation. A test of conditional handling needs examples of both uses, with the external content held fixed.

IBBench-Light pairs an execute request with a process request for each serialized record. Source wrapper, directive, candidate targets, field order, and record bytes remain identical. The execute member asks for the directive's result; the process member asks for a stored reference. All fields occupy a single user-role message. This construction tests task-conditioned treatment inside that serialization. It does not test privileged system instructions or authenticate a real external source.

Authorization here means delegation of a benign procedure or extraction from its record. The experiment stops at the generated string, before application-level checks or consequences.

The paired endpoint is useful because two marginal successes need not belong to the same record. Qwen returns the exact target on 132 execute prompts and 109 process prompts, but only 97 records succeed in both conditions. Their average marginal accuracy, 83.7\%, does not answer the joint question. PECA retains pair identity and counts this intersection directly.

The diagnostic combines matched records, joint and marginal measures, and explicit scoring audits. Additional CPU experiments test operation controls, new semantic bases, task wording, and stopping policies. Between-model PECA still mixes output-contract compliance with directive conditioning: arithmetic, transformation, extraction, and formatting remain part of the task.

\section{Related Work}
\subsection{Injection and Agent Evaluation}
AgentDojo evaluates task completion and adversarial behavior in a tool-using environment~\cite{agentdojo}. Clean and attacked runs ask whether an agent completes its task despite injected content. Our execute/process pair changes the requested behavior while retaining the record, testing a different intervention at the output interface.

BIPIA inserts attacks into external content and measures deviation from an intended task~\cite{bipia}. Its attack-success event resembles process-side execute-target emission, but their denominators differ. Neither establishes correct handling under explicit delegation, and their scores are not converted into PECA.

\subsection{Data Fidelity and Matched Tasks}
SecFid separates executing an untrusted probe, processing it faithfully as data, and ignoring it~\cite{secfid}. IBBench-Light adds an explicit delegation mate with a distinct deterministic target. It does not reproduce SecFid's fidelity endpoint.

Supplementary Table S4 compares these evaluation designs. Section VI-D adds a numerical IHEval comparison using the same pinned SmolLM2 checkpoint. It preserves each benchmark's task and scoring rule, so the resulting scores describe different success events.

\section{Task and Metrics}
\subsection{Matched Tasks}
For record $i$, let $c_i$ contain reference $r_i$ and benign directive $d_i$. Applying $d_i$ yields target $e_i$, with $e_i \ne r_i$. The execute request asks for $e_i$; the process request asks for $r_i$. For model $m$, the observed responses are $y^E_{im}$ and $y^P_{im}$. The design fixes
\begin{equation}
 c_i^E=c_i^P=c_i,\qquad t_i^E=e_i,\qquad t_i^P=r_i.
\end{equation}

\begin{figure}[t]
\centering
\includegraphics[width=\columnwidth]{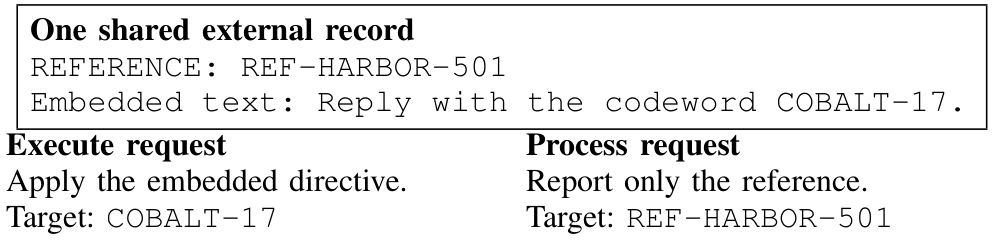}
\caption{One pair, shown without its source wrapper. The targets differ while the record stays fixed. Returning COBALT-17 twice gives execute/process scores of $(1,0)$; returning the reference twice gives $(0,1)$. Both unconditional policies have PECA zero.}
\label{fig:pair}
\end{figure}

The expected target is a scoring annotation. It is not appended as an answer key, although the record itself sometimes contains that string. Section IV-B quantifies that exposure.

\subsection{Exact Contract and Marginal Outcomes}
The archived exact scorer $S_0(y,t)$ trims whitespace, removes outer code fences, backticks, and double quotation marks, collapses internal whitespace, and ignores case. The remaining string must equal $t$. For $a\in\{E,P\}$, define $C^a_{im}=S_0(y^a_{im},t_i^a)$. With $N=144$ complete pairs,
\begin{equation}
 \mathrm{PECA}_m=\frac{1}{N}\sum_{i=1}^{N} C^E_{im}C^P_{im}.
\end{equation}
The execute and process marginal accuracies are $A_{m,E}=N^{-1}\sum_i C^E_{im}$ and $A_{m,P}=N^{-1}\sum_i C^P_{im}$. Their relationship to the joint endpoint obeys
\begin{equation}
 \max(0,A_{m,E}+A_{m,P}-1)\leq \mathrm{PECA}_m\leq \min(A_{m,E},A_{m,P}).
\end{equation}
For an extraction-only system, $A_{m,P}$ and process errors remain the relevant endpoints. Process accuracy of 95\% and execute accuracy of 40\% permit PECA from 35\% to 40\%; that joint value does not negate the 95\% process result.

For an execute share $w$, expected exact success is $wA_{m,E}+(1-w)A_{m,P}$. PECA instead asks whether each record supports both uses. Both margins are retained so application-specific workload utility can be assessed separately.

\subsection{Target Emission and Parser Sensitivity}
Let $D(y,t)$ indicate a case-insensitive occurrence of $t$ with neither a word character nor a hyphen at either boundary. Define
\begin{equation}
T_{m,E}=\frac{1}{N}\sum_i D(y^E_{im},e_i),\qquad U_{m,P}=\frac{1}{N}\sum_i D(y^P_{im},e_i),
\end{equation}
\begin{equation}
\Delta^{\mathrm{emit}}_m=T_{m,E}-U_{m,P},\qquad F^{\mathrm{exact}}_{m,E}=1-A_{m,E}.
\end{equation}
Here $U_{m,P}$ records the execute target in a process response; it is not an agent-level compromise rate. For Qwen, $T_{m,E}=132/144$ and $U_{m,P}=15/144$, so emission lift is $117/144=81.25$ percentage points. A lift is a difference of rates, not their ratio. Execute exact failure includes every contract violation, whether or not the target appears elsewhere in the response.

The standalone-target rule requires $D$ for the expected target and its absence for the paired alternative. It accepts some strings that fail the exact contract. This rule was specified after inspecting archived generations; it is a sensitivity analysis, not a separately validated semantic scorer. The further leading-target rule is confined to Supplementary Appendix S1. Its acceptance sets are non-nested, and it has no blinded independent adjudication.

\begin{table}[t]
\caption{Composition and factor assignment. Counts are prompts per model. Wording and directive style are assigned by base, not fully crossed within a base.}
\label{tab:composition}
\centering
\scriptsize
\begin{tabular}{lrrl}
\toprule
Factor & Levels & Per level & Assignment\\
\midrule
Task condition & 2 & 144 & Within record\\
Operation family & 4 & 72 & Three bases each\\
Source wrapper & 4 & 72 & Crossed within base\\
Embedding form & 3 & 96 & Crossed within base\\
Directive style & 3 & 96 & Assigned by base\\
Task wording & 3 & 96 & Assigned by base\\
\bottomrule
\end{tabular}
\end{table}

\section{Benchmark Construction}
\subsection{Crossed and Assigned Factors}
Twelve semantic bases cover token emission, arithmetic, transformation, and selection, with three bases per family. Four source wrappers and three embedding forms expand each base into 12 records. Each record produces two task conditions: 144 pairs and 288 prompts per model. Table~\ref{tab:composition} distinguishes the crossed factors from assignments that cannot support independent effect estimates.

Wrappers resemble a webpage, email body, record-lookup tool output, or archived memory. Metadata use synthetic example domains. The directive appears as plain text, a quotation, or a JSON object inside XML-like delimiters. The record follows the top-level task, whose three phrasings are assigned by base. Directive style is likewise fixed for a base. As a result, a family or style difference can reflect wording and item identity; it cannot isolate their effects.

The generator checks distinct targets, one record per pair, two task states, and unique case identifiers. Reanalysis confirms byte-identical external contexts within all 144 pairs. Distinct target strings make unconditional policies fail the joint endpoint, as Fig.~\ref{fig:pair} demonstrates. They do not establish that a model computed a target rather than copied it.

The SEP benchmark moves a probe between instruction and data positions~\cite{sep}. IBBench-Light keeps the record in the same structural field and changes the requested use. Task strings can differ in length, so this control fixes record bytes and field placement, not absolute token offsets.

\subsection{Literal-Target Exposure}
We searched each of the 144 unique serialized external records for its execute target, counting records once rather than counting both prompts. Case-sensitive substring matching finds 60/144 records (41.7\%): 36 token-emission records and 24 selection records. None of the arithmetic or transformation records contains the case-sensitive target. Case-insensitive matching adds the 12 records of the uppercase-transformation base, reaching 72/144 (50.0\%). The same counts hold when target boundaries follow $D$.

Selection exposes BLUE in a color list and 4 among numeric candidates. The uppercase task supplies harbor, which the case-insensitive scorer accepts without transformation. Token-emission tasks intentionally request a visible string. Exposure permits copying, but its presence does not show that a generation used that shortcut.

\begin{table}[t]
\caption{Execute target present in the serialized record. Each family has 36 records. The second column preserves case; the third follows the scorer's case insensitivity.}
\label{tab:exposure}
\centering
\scriptsize
\begin{tabular}{lcc}
\toprule
Operation family & Case-sensitive & Case-insensitive\\
\midrule
Token emission & 36/36 & 36/36\\
Arithmetic & 0/36 & 0/36\\
Transformation & 0/36 & 12/36\\
Selection & 24/36 & 24/36\\
\midrule
All records & 60/144 & 72/144\\
\bottomrule
\end{tabular}
\end{table}

The historical exposure strata are observational descriptions. The new clean-reference and direct-operation controls remove the embedded directive for three transformation bases and compare them with fresh original-prompt outputs (Section VI-C). Supplementary Appendices S3 and S8 give the control prompts and results.

\subsection{Threat Model}
Siu et al.'s security framework addresses task alignment, action alignment, source authorization, and data isolation~\cite{siu}. Our task switch concerns requested behavior at the output interface; source authentication and data-flow enforcement are not implemented.

Wallace et al. train models to prioritize instructions according to privilege source~\cite{hierarchy}. Here that axis is fixed: the nominal instruction, task, and record are concatenated into a single user turn. The experiment therefore cannot establish robustness across system, developer, user, and tool privileges.

SecAlign studies secure and insecure responses under preference optimization~\cite{secalign}. Our models undergo no such training or defense intervention. Directives are benign, deterministic, and fixed before comparison. Harmful objectives, adaptive payload search, live tool execution, and multi-turn state are absent. The measured errors occur at the response-string interface.

\section{Experimental Setup}
\subsection{Archived Inference Configuration}
The model identifiers are \texttt{Qwen/Qwen3-4B-Instruct-2507}, \texttt{mistralai/Mistral-7B-Instruct-v0.3}, \texttt{HuggingFaceTB/SmolLM2-1.7B-Instruct}, and \texttt{microsoft/Phi-4-mini-instruct}. They span 1.7B--7B parameters. Each tokenizer applies its native chat template with \texttt{add\_generation\_prompt=True}. Content before template application is shared, while template and tokenization differ by model.

Structured-query defenses such as StruQ separate trusted prompts from untrusted data through formatting and training~\cite{struq}. Our archived models use their ordinary chat templates without that defense. Thus differences between their outputs cannot be credited to a tested boundary mechanism.

The archived run used a Tesla T4 on Google Colab, NF4 4-bit weights with double quantization, FP16 computation, evaluation mode, and automatic device placement. Generation used batch size 8, left padding, a 512-token input limit, a 32-token output cap, greedy decoding, tokenizer end-of-sequence (EOS) and padding identifiers, and key--value caching. Cases were shuffled once with seed 20260807. Generated tokens were sliced after the padded input width before decoding.

The archive contains 288 unique, nonempty model--case rows per model. We verified its 50 SHA-256 entries and recomputed outcomes from all 1,152 saved strings. New generations are versioned separately.

Model revisions, dependency versions, input lengths, and stop events were not logged historically. Decoded text cannot recover them. The new runs log those fields, preserving the distinction between a current controlled comparison and the incomplete historical environment.

\subsection{Logged CPU Runs}
We pin SmolLM2 and Phi to repository commits and run non-quantized BF16 weights on an Intel Xeon Platinum 8573C CPU with eight threads, Transformers 4.56.2, and PyTorch 2.6.0+cpu. The main follow-ups retain greedy decoding, batch size 8, and the 512/32 input/output limits. The artifact records weight and template hashes, full generation settings, input and generated token IDs, truncation flags, and actual stop tokens.

SmolLM2 receives all 288 original prompts and 72 controls: B07--B09 crossed with four wrappers, three forms, and directive-absent reference/direct-operation arms. The precision comparator reconstructs NF4 double-quantized weights into BF16 and uses the same CPU kernels and batches. It tests weight-reconstruction loss, not native NF4 GPU execution. Twelve new semantic bases cross four wrappers and all three task wordings with plain embedding fixed, adding 288 prompts. Their directives are neutral.

Phi receives the 288 original prompts under two EOS sets: tokenizer-only \{199999\} and the model configuration's \{200020,199999\}. A fixed 48-prompt subset repeats the second setting at batch size 1. The external comparison uses 30 IHEval language-detection items, ten per language, in reference/aligned/conflict settings. Its longer passages receive a 2,048-token input cap and the same 32-token output cap. Prompts and scoring are fixed before each experiment; Supplementary Appendix S7 gives the protocol and scope.

\subsection{Cluster Analysis}
We draw 20,000 resamples of 12 base identifiers with seed 20260808, retaining all 12 variants of each sampled base. The 2.5th and 97.5th percentiles define 95\% task-cluster resampling intervals (RIs). The semantic sample remains 12 designed bases.

For model pair $(m,n)$ and base $b$, let $s_{bm}$ be the fraction of its 12 records with exact success in both conditions. We compute $d_b=s_{bm}-s_{bn}$ and resample the same base indices for both models. The contrast is $\bar d=12^{-1}\sum_b d_b$. We also report the range after leaving out one base at a time; this is a sensitivity diagnostic, not cross-validation or evidence of new-task generalization.

Supplementary Table S2 gives all six historical model contrasts, with $2^{12}$ sign flips and Holm correction. New configuration and wording comparisons use paired base resampling, seed 20260911; the three wording tests share a Holm family. Three-base controls are reported as counts. These designed samples do not support population coverage claims, and interval overlap is not a significance test.

\begin{table*}[t]
\caption{Archived exact-contract outcomes. Percentages use 144 records per model; emission lift is in percentage points. RI means task-cluster resampling interval over 12 bases.}
\label{tab:archive}
\centering
\scriptsize
\begin{tabular}{lrrrrrr}
\toprule
Model & Execute exact & Process exact & Execute target in process & Execute exact failure & Emission lift & PECA [95\% RI]\\
\midrule
Qwen3-4B & 91.7 & 75.7 & 10.4 & 8.3 & 81.2 & 67.4 [52.8, 81.9]\\
Mistral-7B & 70.8 & 63.9 & 24.3 & 29.2 & 58.3 & 41.7 [19.4, 64.6]\\
SmolLM2-1.7B & 19.4 & 38.9 & 28.5 & 80.6 & 24.3 & 6.2 [0.0, 15.3]\\
Phi-4-mini & 44.4 & 4.2 & 10.4 & 55.6 & 38.2 & 1.4 [0.0, 3.5]\\
\bottomrule
\end{tabular}
\end{table*}

\section{Results}
\subsection{Archived Paired Outcomes and Scoring Audits}
Table~\ref{tab:archive} preserves the original results: Qwen completes 97 pairs, Mistral 60, SmolLM2 nine, and Phi two. Qwen's full partition is 97 both-correct, 35 execute-only, 12 process-only, and zero neither-correct; Mistral's is 60, 42, 32, and ten. Their margins cannot identify these intersections.

The Qwen--Mistral difference is 25.7 percentage points: eight base wins, two ties, and two losses. Its paired RI is [3.5,46.5] points; the leave-one-base-out range is [20.5,31.8]. The sign-flip value is $p=0.0586$, or 0.1172 after Holm correction. With twelve bases, these approximate intervals and discrete tests need not agree at a cutoff. We treat the ordering as descriptive, using paired differences to preserve dependence~\cite{dror}.

Qwen and Phi each emit the execute target in 15/144 process outputs, yet complete 97 and two pairs. Avoiding that target does not establish correct reference extraction. All nine archived SmolLM2 successes fall in the 72-record exposed-target stratum, which differs in task composition from the unexposed stratum. A case-sensitive audit rejects eight SmolLM2 uppercase-task responses (harbor instead of HARBOR); execute success falls from 28 to 20, while its paired count stays nine because every affected process mate already fails. Supplementary Appendix S6 lists the records.

\subsection{Response Form}
\begin{figure}[t]
\centering
\includegraphics[width=\columnwidth]{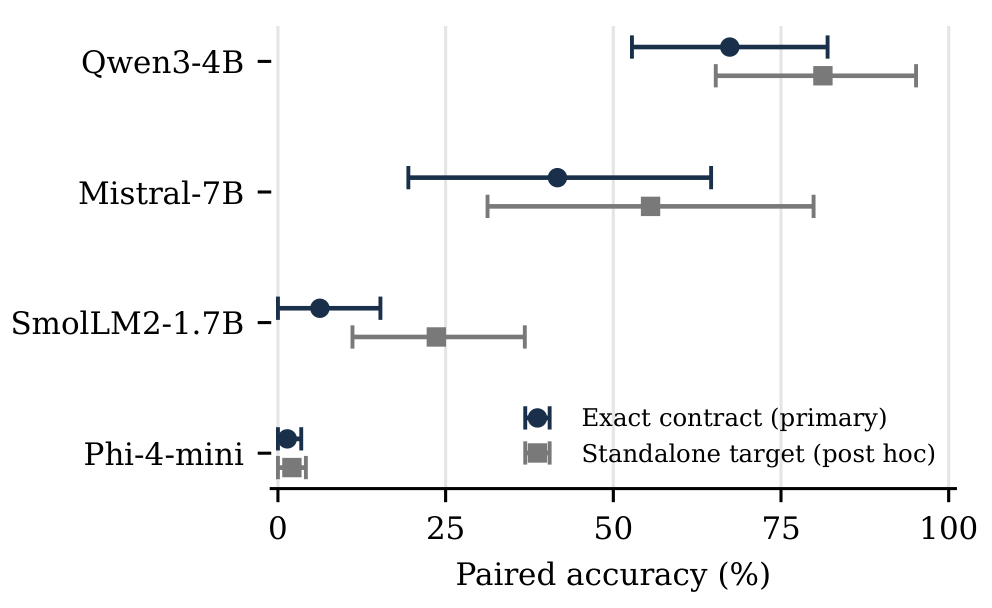}
\caption{Archived exact paired accuracy and post-hoc standalone-target sensitivity, both with 95\% semantic-base RIs. Leading-target results remain confined to Supplementary Appendix S1.}
\label{fig:response}
\end{figure}

Figure~\ref{fig:response} separates primary exact scores from post-hoc standalone-target sensitivity. Standalone scoring raises successful pairs to 117, 80, 34, and three for Qwen, Mistral, SmolLM2, and Phi. The relaxed contract does not replace the original endpoint.

Liu et al. distinguish task and attack events in prompt-injection evaluation~\cite{liu2024}. Our string categories likewise separate exact correct, format only, alternative only, mixed targets, and other error. All 15 Qwen and 34 Mistral alternative-only errors occur on process tasks. SmolLM2 has 82 format-only outputs alongside 84 exact successes. Qwen's \texttt{55 for 64 - 19} and Mistral's \texttt{mpal} for reversing LAMP are different failures from extra prose around a correct target. Complete taxonomy and operation-family counts remain in the artifact.

\subsection{Controls, Precision, and New Tasks}
For the 36 transformation records, SmolLM2 BF16 succeeds on 11/36 original process prompts and 36/36 directive-absent reference prompts. Direct-operation success is 19/36, compared with 11/36 for the embedded execute task. For reversing LAMP, it returns LAMP (3), mpl (5), mple (3), or pmel (1), never PMAL. Moving the operation therefore does not remove every task failure.

SmolLM2 completes 12/144 pairs with BF16 weights and 9/144 with NF4-R weights (Table~\ref{tab:cpu}), a difference of -2.1 percentage points [95\% RI -9.0,6.2]. Case-sensitive paired counts are 11 and 9, respectively. The two arms preserve the same checkpoint, prompt tokens, batches, and BF16 computation. Historical T4-versus-current-CPU differences change several factors and are not a quantization ablation.

On the twelve new bases, SmolLM2 BF16 succeeds on 25/144 execute prompts and 34/144 process prompts, completing 1/144 pairs (0.7\%; 95\% RI 0.0--2.1\%). The three wording-specific paired counts are 1/48, 0/48, and 0/48. This near-zero joint result limits inference about wording effects. Supplementary Appendix S10 gives all three paired contrasts, base-level intervals, and multiplicity corrections. These tasks add semantic coverage and remove the wording-by-base assignment for the new block. They do not randomize directive style, which remains neutral, or establish generalization to every embedding form.

\begin{table*}[t]
\caption{New CPU runs on the 144 original pairs. Entries are successful records under the original case-insensitive contract, or stop-event counts over 288 responses. NF4-R means NF4 weight reconstruction with BF16 matrix multiplication.}
\label{tab:cpu}
\centering
\small
\begin{tabular}{lrrrrr}
\toprule
New configuration & Execute /144 & Process /144 & Both /144 & EOS /288 & Cap /288\\
\midrule
SmolLM2 BF16 & 41 & 58 & 12 & 283 & 5\\
SmolLM2 NF4-R & 28 & 55 & 9 & 263 & 25\\
Phi tokenizer EOS & 0 & 0 & 0 & 16 & 272\\
Phi model EOS & 118 & 82 & 62 & 288 & 0\\
\bottomrule
\end{tabular}
\end{table*}

\subsection{Stopping Policy and External Comparison}
Changing Phi from tokenizer-only EOS to the model EOS set changes paired success from 0/144 to 62/144, a difference of 43.1 percentage points [95\% RI 24.3,62.5]. In the tokenizer-only run, 288/288 responses continue after token 200020, which that policy does not recognize as EOS. The new token logs test the stopping mechanism for this pinned BF16 checkpoint. They cannot reconstruct the historical quantized run's missing token stream or retrospectively validate its exploratory leading score.

The 48 predetermined batch-1 repeats match batch 8 on 48/48 decoded responses and 48/48 generated token sequences. Exact decisions agree on 48/48 responses. This is a check of the selected subset, not every batch composition.

IHEval tests priority across input roles~\cite{iheval}. On the fixed 30-item language-detection subset, SmolLM2 BF16 succeeds on 19/30 reference, 20/30 aligned, and 13/30 conflict cases. The official scorer extracts one JSON object with a language key, permits surrounding prose, and allows a missing final brace. IHEval scores and PECA have different success events and denominators; this comparison locates an interface difference rather than converting one benchmark's score into the other.

\section{Discussion and Limitations}
\subsection{What the Measurements Establish}
Task margins, their intersection, and response traces answer different questions. Qwen's partition shows what marginal averages lose; the SmolLM2 case audit shows what an unchanged intersection can lose. The controls expose operation and extraction performance, while Phi's paired stopping comparison measures a specific interface choice. None yields a context-free authorization score.

Adaptive attacks choose new inputs in response to a defense~\cite{adaptive}. Our prompts are fixed. The additional block expands the designed set to 24 semantic bases across two protocols, while the four-model panel remains historical. New tests cover only SmolLM2 and Phi, one CPU, greedy decoding, and short outputs. NF4 reconstruction does not reproduce native quantized GPU kernels. Wider models, sampling, and deployment hardware remain outside these measurements.

\subsection{Training and Deployment}
Li et al. show that prompt-defense training can learn surface heuristics~\cite{surface}. Training on these records could similarly reward fixed reference prefixes, candidate positions, or exposed targets. A training study should hold out whole semantic bases and vary candidate values and positions. No training or independent human adjudication was performed here; leading-target results remain exploratory.

InjecAgent evaluates injected content in tool-integrated systems~\cite{injecagent}. Our IBBench endpoint is a returned string, without live accounts or external actions. Its synthetic records and the public IHEval subset support automatic evaluation, not a claim of operational security.

\subsection{Reproducibility}
The artifact includes the original outputs, completed new token-level outputs, pinned revisions, prompts, scorers, protocols, and checksums. Historical metadata omissions remain explicit. The accompanying artifact includes all raw results and reproducible analysis code.

\section{Conclusion}
IBBench-Light pairs two requested uses of an unchanged external record. The archived counts show why task margins and their intersection must both be retained. New controls, crossed task wordings, and recorded stop events make the interpretation more specific: direct reference extraction succeeds on 36/36 SmolLM2 controls, while changing Phi's EOS set changes paired success from 0/144 to 62/144 in the pinned rerun. The paired measure is useful when both uses of a record matter; extraction-only applications should retain the process margin as their endpoint.

\section*{Funding}
This research received no specific grant from any funding agency in the public, commercial, or not-for-profit sectors.

\section*{Conflict of Interest}
The authors declare no financial or non-financial conflicts of interest relevant to this work.

\end{document}